%% file: sidtd_paper.tex
\documentclass{article}

\usepackage{PRIMEarxiv}

\usepackage[utf8]{inputenc} 
\usepackage[T1]{fontenc}    
\usepackage{hyperref}       
\usepackage{url}            
\usepackage{booktabs}       
\usepackage{amsfonts}       
\usepackage{nicefrac}       
\usepackage{microtype}      
\usepackage{lipsum}
\usepackage{graphicx}
\graphicspath{{media/}}     

\usepackage{lineno}
\usepackage{listings}
\usepackage{subcaption}
\usepackage{adjustbox}
\usepackage{multirow}
\usepackage{xcolor}

\title{Synthetic dataset of ID and Travel Documents\footnote{This content reflects only the authors' view. The European Agency is not responsible for any use that may be made of the information it contains. }
\thanks{\textit{\underline{Citation}}: 
\textbf{C. Boned, M. Talarmain, N. Ghanmi, G. Chiron, S. Biswas, A. M. Awal, O. Ramos Terrades. Synthetic dataset of ID and Travel Documents. }} 
}

\author{
  C. Boned, M. Talarmain,  S. Biswas \\
  Computer Vision Center \\
   Bellaterra, 08193, Spain \\
  \texttt{\{cboned, mtalarmain, sbiswas\}@cvc.uab.cat} \\
   \And
  N. Ghanmi, G. Chiron,  A. M. Awal \\
  IDNow, \\
  AI \& ML Center of Excellence, 35220, France \\
  \texttt{\{nabil.ghanmi,montaser.awal\}@idnow.io} \\
   \AND
    O. Ramos Terrades\footnote{Corresponding author} \\
  Computer Vision Center \\
  Department of Computer Science \\
  Universitat Autònoma de Barcelona \\
   Bellaterra, 08193, Spain \\
  \texttt{oriolrt@cvc.uab.cat} \\
}

\begin{document}
\maketitle

\begin{abstract}
This paper presents a new synthetic dataset of ID and travel documents, called SIDTD. 
The SIDTD dataset is created to help training and evaluating forged ID documents detection systems.
Such a dataset has become a necessity as ID documents contain personal information and a public dataset of real documents can not be released.
Moreover, forged documents are scarce, compared to legit ones, and the way they are generated varies from one fraudster to another resulting in a class of high intra-variability.
In this paper we trained state-of-the-art models on this dataset and we compare them to the performance achieved in larger, but private, datasets. 
The creation of this dataset will help to document image analysis community to progress in the task of ID document verification.
\end{abstract}

\keywords{Dataset \and Forgery detection \and ID Documents verification}

\section*{Background \& Summary}

The development of remote identity authentication systems, which include both biometrics and ID and travel documents verification, has increased and spread since the advent of the COVID-19 pandemic. These authentication systems have allowed people to work and to develop their business activities out of their offices as public administration, banks, and productive industries and many services have adopted them in their usual workflow.  These services offer an online enrollment, thus, avoiding the user's physical attendance by requiring a selfie and a picture of their ID document to authenticate them. However, cybercrime has taken advantage of society’s vulnerabilities and evolve towards more sophisticated threats. As pointed by the IOCTA 2020 report\cite{IOCTA2020}, ``the fundamentals of cybercrime are firmly rooted, but that does not mean cybercrime stands still. Its evolution becomes apparent on closer inspection, in the ways seasoned cybercriminals refine their methods and make their artisanship accessible to others through crime as a service''. In particular, fraudsters may take advantage of these vulnerabilities by forging ID documents to alter information or hide their real identity. Consequently, new developments on identity authentication systems must include advanced AI tools to reliably ensure citizen’s security and protect the online services.

A key tool to ensure citizen's identity on a digital environment, among others, is the detection of forged ID and travel documents when they enrol to online services. This {\em Presentation Attack Detection} (PAD) tool must compare an image, or video, most likely acquired by mobile devices, of citizen's ID documents and  to asses if such ID document images, or videos, corresponds to a {\em bona fide} document or not. Given the current legislation on data protection, as the GDPR in the EU, publish real ID documents data   is restricted to those in which citizens has provide explicit consent. Consequently, it is difficult to gather enough data to estimate model parameters  to detect forged documents  and it has been developed sophisticated AI models that generate synthetic ID and travel  Documents~\cite{Bulatov2021}. These models train Generative Adversarial Networks (GANs) to simulate ID Documents containing information from not existing people. Despite being models that generate quite realistic ID Documents, the generated data is useless for PAD tasks as they do not contain the security features that document of this kind usually are added them.

The current trend is therefore to detect ID Documents that have been altered by detecting unexpected changes on the document texture, text or identity photo location. Thus, GAN-based models are proposed to generate ID document images from a limited set of templates under three typical {\em presentation attack instruments}\cite{Benalcazar2022} (PAI), namely: {\em composite}, {\em print} and {\em screen}, as they are defined by the  International Standard (ISO/IEC 30107)\cite{iso3017} and  performance evaluation  of general purpose classification networks on two tasks: {\em composition detection} and {\em source detection} is  done. The first task aims at detecting the {\em composite} PAI while the second task aims at detecting whether the ID Document comes from a {\em bona fide} document or a {\em printed} or a {\em screened} copy of a {\em bona fide} document.  In any case, synthetic datasets used on that experiments are not published and cannot be actually used for benchmarking purposes.

In this paper we introduce an extension  of the MIDV2020 dataset\cite{Bulatov2021} for PAD purposes given the above-mentioned PAI tasks. The MIDV2020 Dataset is the latest version of the MIDV identity document dataset series, which is the largest publicly available identity documents dataset with variable artificially generated data. The proposed dataset, the SIDTD dataset, contains the original MIDV2020 images and videos, which will compose the corpus of {\em bona fide} documents, together with a set of images and videos, that are the altered version of the MIDV2020 ID Documents and  compose the corpus of forged documents.

The SIDTD dataset, as the MIDV2020 dataset, is licensed under a \hyperlink{https://creativecommons.org/licenses/by-sa/2.5/}{Generic Creative Commons Attribution-ShareAlike 2.5} and it is fully  stored in the CVC dataset repository.  Moreover, we also publish the Python code to download and to use the dataset together with the trained models described in the Technical Validation Section. Finally, the code used to generate the altered ID Documents and to train the models are also available at the same public repository: \hyperlink{https://github.com/Oriolrt/SIDTD\_Dataset}{https://github.com/Oriolrt/SIDTD\_Dataset}.

\section*{Methods}\label{sec:methods}
As explained above, the SIDTD dataset is an extension of the MIDV2020 dataset~\cite{Bulatov2021}. Initially, the MIDV2020 dataset is composed of forged ID documents, as all documents are generated by means of AI techniques. These generated documents are considered in the SIDTD dataset as representative of {\em bona fide}.
On the other hand, the documents generated as described in this section will be considered as being {\em forged} versions of them. The corpus of the dataset is composed by ten European nationalities that are equally represented: Albanian, Azerbaijani, Estonian, Finnish, Greek, Lithuanian, Russian, Serbian, Slovakian, and Spanish.

\subsection*{Forged ID Document images generation}

\begin{figure}[ht]
\begin{center}
 \includegraphics[width=0.7\columnwidth]{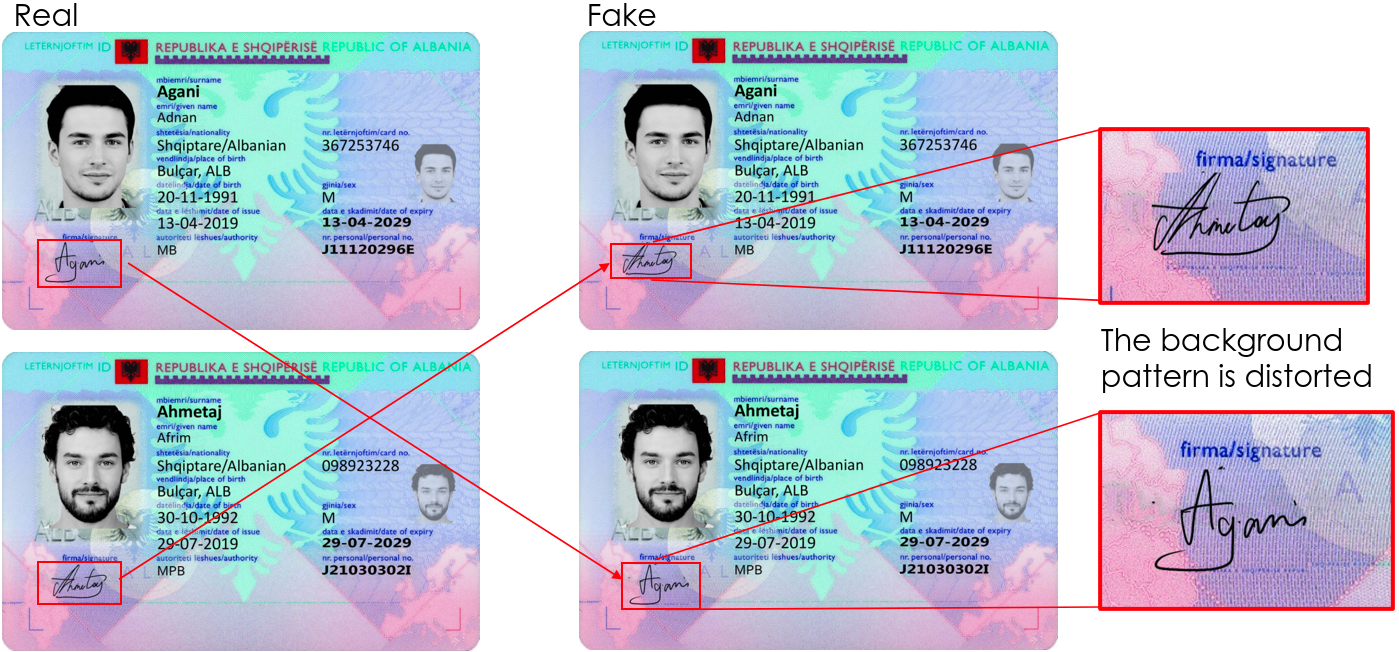}
\end{center}
\caption{Crop \& Replace Composite PAI example. The signature of two ID documents of the same nationality is replaced.}\label{fig:crop_replace}
\end{figure}

We employ two techniques for generating \emph{composite} PAIs: \emph{Crop \& Replace} and \emph{inpainting}. The \emph{Crop \& Replace} technique is a fundamental image processing approach that involves the exchange of information between two identification (ID) documents of the same class. This is achieved by cropping a specific region from one ID document and replacing it with corresponding information from another ID document, as illustrated in \figurename~\ref{fig:crop_replace}. To mitigate the risk of creating a perfect match and ensure the artificial documents are indistinguishable from their authentic counterparts, we introduce a \emph{shift} parameter. This parameter determines the offset for the exchanged regions. Thus, we define a range $[-n, n] \setminus \{0\}$, where $n \in \mathbb{N}$, for the random setting of shift parameters along both the x-axis and y-axis. The shift value $0$ is excluded to prevent perfect matching. The introduced shift parameter induces a border effect resulting from texture discontinuity, which must be detected by the PAD method.

Conversely, the \emph{Inpainting} technique is a sophisticated image processing method that involves replacing a small region of an image while maintaining a realistic look and feel. This technique is commonly applied in post-production to remove people from pictures or architectural artifacts from images and movies. In the context of ID documents, {\em Inpainting} can be applied to eliminate personal information, such as names or surnames, from textual fields, replacing it with false information of the same nature.

\begin{figure}[ht]
\begin{center}
 \includegraphics[width=0.7\columnwidth]{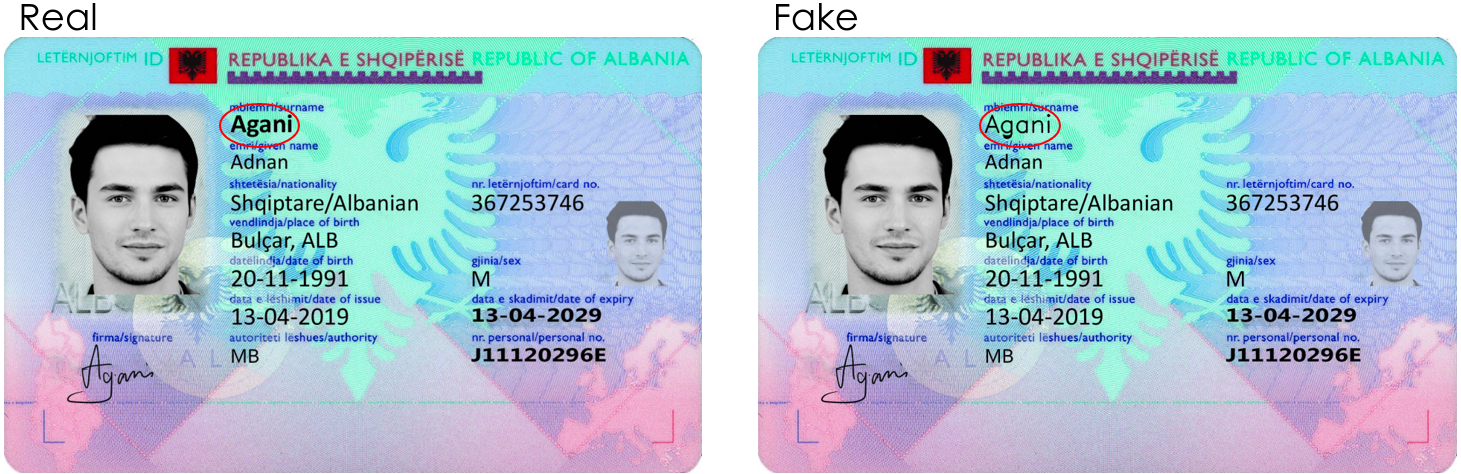}
\end{center}
\caption{Inpainting Composite PAI example. The name field in the ID document is replaced by the same content but changing the font.}\label{fig:inpainting}
\end{figure}

First, {\em Inpainting} is employed to remove the original information by generating a realistic background that covers the text. Then, the size of the newly added text is computed through interpolation and inference based on the information surrounding the text fields, and the font is randomly selected from a set of available font types. An illustrative example of a forged ID document is presented in \figurename~\ref{fig:inpainting}, where the name was regenerated using the {\em Inpainting} technique.  

Depending on the shift value in the {\em  Crop \& Replace } technique and the chosen fonts and text sizes for each manipulation, the appearance of the generated ID documents can range from easily distinguishable to humans to extremely subtle and indistinguishable alterations.

\subsection*{Forged ID Document videos generation}

\begin{figure}[ht]
    \centering
    \includegraphics[height=45mm]{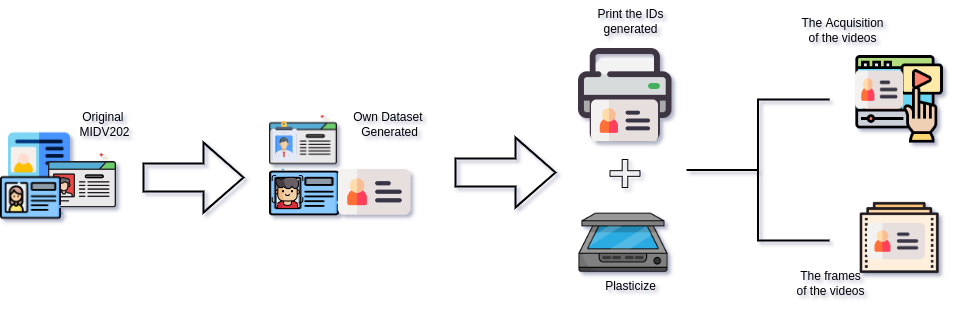}
    \caption{Pathflow followed to generate forged videos and clips}
    \label{fig:Vgen}
\end{figure}

As the original MIDV2020 dataset~\cite{Bulatov2021} contains videos, and clips, of captured ID Documents with different backgrounds, we add the same type of data for the forged ID Document images generated using the techniques described in the previous section. The protocol employed to generate the dataset is as follows: We printed 191 counterfeit ID documents, created using the tools detailed in the previous section, on paper using an HP Color LaserJet E65050 printer. Then, the documents were laminated with 100-micron-thick laminating pouches to enhance realism and manually cropped, as depicted in \figurename~\ref{fig:Vgen}. 

\begin{figure}
\captionsetup[subfigure]{justification=centering}%
\begin{subfigure}[t]{.49\textwidth}
  \centering
    \includegraphics[height=5cm]{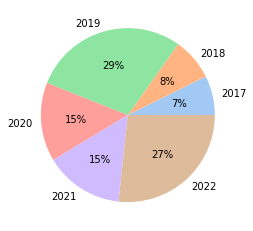}
  \caption{}
  \label{fig:pie_chart_vid_by_year}
\end{subfigure}%
\begin{subfigure}[t]{.49\textwidth}
  \centering
    \includegraphics[height=5cm]{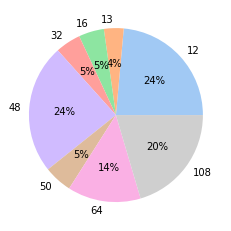}
  \caption{}
  \label{fig:pie_chart_vid_by_cam_quality}
\end{subfigure}%
\caption{Distribution of videos according to: a) phone year of release b) resolution of the smartphone main rear camera (in megapixels)}
\end{figure}

CVC's employees were requested to use their smartphones for recording videos of forged ID documents from SIDTD. This approach aimed to capture a diverse range of video qualities, backgrounds, durations, and light intensities. The resulting dataset includes videos from various smartphone brands such as Samsung (Galaxy A5, Galaxy A52, Galaxy A53 5G, Galaxy A70, Galaxy M12, Galaxy M21, Galaxy M31s, Galaxy S10e, Galaxy S20+ 5G, Galaxy S21, Galaxy S22+), Xiaomi (Mi 10T Pro 5G, Mi 8, Mi 9 Lite, Mi A3, Mi Max 2, POCO M2, POCO X3 Pro, Redmi Note 7 pro, Redmi Note Pro 11+), OnePlus (OnePlus 7 Pro, OnePlus 6T), Apple (iPhone 13, iPhone 12, iPhone 11, iPhone 8), Motorola (Moto G12, Moto G31), Google (Pixel 4a), Oppo (Realme C2), each offering a broad spectrum of camera properties (see Figure \ref{fig:pie_chart_vid_by_cam_quality} for the distribution of camera resolutions in megapixels).

\begin{figure}[h]
\captionsetup[subfigure]{justification=centering}%
\begin{subfigure}[t]{.24\textwidth}
  \centering
    \includegraphics[width=.99\textwidth,height=5cm]{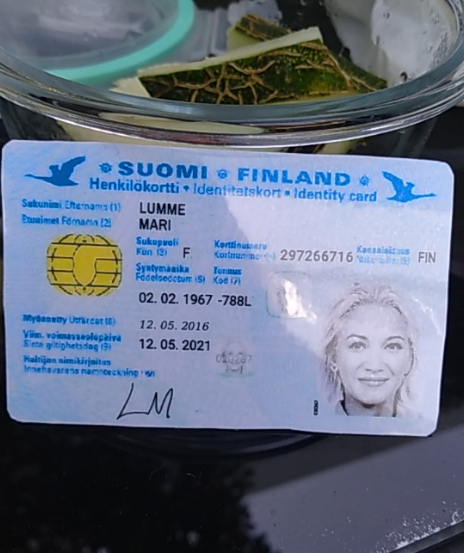}
  \caption*{(a) Finnish ID document}
  \label{fig:sfig1}
\end{subfigure}%
\begin{subfigure}[t]{.24\textwidth}
  \centering
    \includegraphics[width=.99\textwidth,height=5cm]{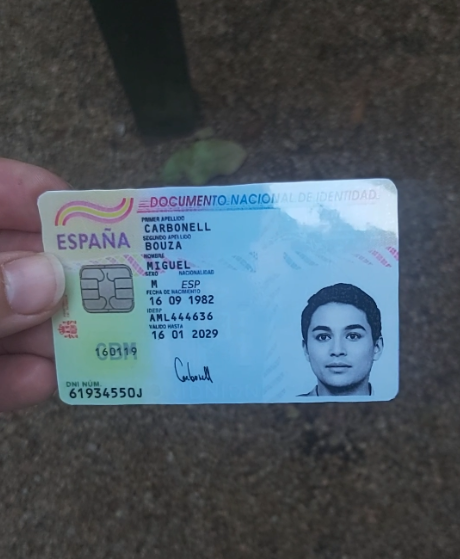}
  \caption*{(b) Spanish ID document}
  \label{fig:sfig2}
\end{subfigure}%
\begin{subfigure}[t]{.24\textwidth}
  \centering
    \includegraphics[width=.99\textwidth,height=5cm]{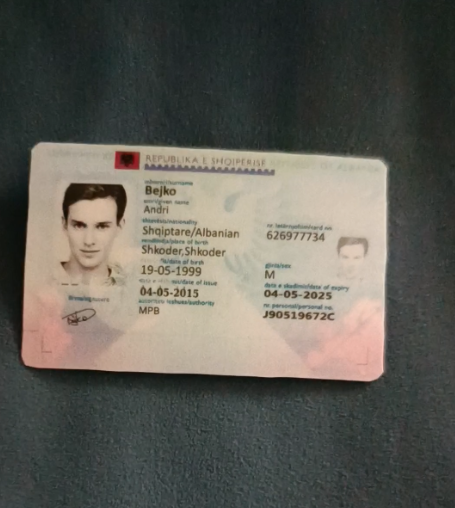}
  \caption*{(c) Albanian  ID document}
  \label{fig:sfig4}
\end{subfigure}
\begin{subfigure}[t]{.24\textwidth}
  \centering
    \includegraphics[width=.99\textwidth,height=5cm]{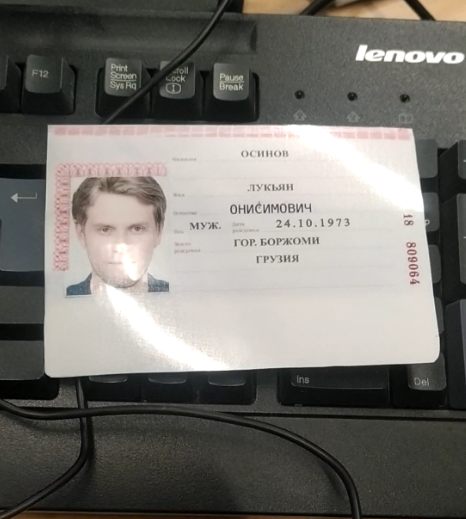}
  \caption*{(d) Russian passport}
  \label{fig:sfig3}
\end{subfigure}
\caption{Examples of SIDTD video clips of fake documents with different backgrounds, lightening and devices: a) Natural light with table background recorded with Xiaomi Mi Max 2 b) Natural light with outside floor background recorded with Samsung Galaxy A70 c) low lighting with chair background recorded with Xiaomi Redmi Note Pro 11+ d) artificial indoor light with keyboard background recorded with Xiaomi Mi A3.}
\label{fig:ex_clips_fake_doc}
\end{figure}

The recorded videos have relatively short durations, ranging between 4s and 13s, with an average of 7s. This duration is similar to {\em bona fide} ID document videos, which also average around 7s, varying between 4s and 12s. Overall, this procedure not only ensures diversity in the dataset but also enriches it with a variety of camera properties and conditions for robust model training. Most of the videos (approximately 85\%) were captured using smartphones released within the last four  years, as depicted in \figurename~\ref{fig:pie_chart_vid_by_year}. Also,  image quality does not only depend on the resolution of the smartphone's primary rear camera\footnote{Source: https://petapixel.com/2022/02/10/image-quality-is-more-than-megapixels/} but several other parameters, such as image enhancement (improving brightness, contrast, colors, and reducing noise) and sensors features (including the number of sensors, sensor size, and sensor quality) play crucial roles in determining overall image quality. As these parameters vary from one smartphone to another, we cannot directly infer the image quality from the information showed in the \figurename~\ref{fig:pie_chart_vid_by_cam_quality}, we can thus note that the images in our dataset have been captured with a wide variety of devices using different image enhancement methods. Despite  we gave the  same instructions to each person, the angles, the movement, the video duration, the position of the document vary from one person to another. It causes  more variability to the dataset and it results in a high diversity  that will help to reduce model overfitting, see \figurename~\ref{fig:ex_clips_fake_doc}.

Finally, we extracted video clips from the recorded videos, every 6 frames as it was done for the MIDV2020 dataset. As a result, a total of 7,214 frames were collected. We annotated each corner of the identity document automatically using SmartDoc 2017's video capture method\cite{chazalon2017smartdoc} based on the open source code they published on Github\footnote{https://github.com/smartdoc2017-competition/dataset\_creation}. The annotations are provided in JSON format with the same annotation structure as the one made for the MIDV2020 dataset.

\begin{table}[ht]
    \centering
    \begin{tabular}{cc|c|c}
    &     &  MIDV2020 & Generated\\ \hline
    \multirow{3}{*}{source} &    template & 1,000  & 1,222 \\
        & video &  1,000 &  191 \\
        & clip & 68,409 & 7,214\\ \hline 
    \end{tabular}
    \caption{Statistics of the SIDTD dataset.}
    \label{tab:sidtd_statistics}
\end{table}

\section*{Data Records}

The SIDTD dataset contains ID documents  in three different formats: templates, videos, and clips. Each format includes both {\em bona fide} and {\em forged} documents. The bona fide documents are the templates, videos and clips released with the MIDV2020 dataset and they consist of 1,000 unique templates. According to the MIDV2020  authors, each document and its own content has been generated as follows: unique faces have been artificially generated by the Generated Photos service\footnote{https://generated.photos/} via a StyleGAN~\cite{karras2019style} model. Each of them is  associated to unique text fields whose values are generated to match the photo in terms of gender and age. The text font  was chosen to be as close as possible to the corresponding legally issued ID document and the signature were generated to fit the last name written in the document. Then, templates were printed on glossy photo paper, laminated and cropped. Video clips, scanned images and photos were all annotated in JSON format readable with VGG Image Annotator v2.0.11 by the MIDV2020 dataset authors. The forged instances of the SIDTD have been generated as described in the previous section and it consists of 1,222 forged templates and 191 videos of forged documents.  Clips are video frames, sampled every 6 frames. In total, we have extracted 7,214 clips from fake documents and 68,409 from real documents, see Table~\ref{tab:sidtd_statistics}.

For each generated document, a { JSON} file  with  information related to the document generation process, has been created. For MIDV2020 instances, we keep the JSON files structure to preserve MIDV2020 dataset consistency in the SIDTD dataset.  The following code illustrates an example for  one image and one of its regions:
 
 \begin{lstlisting}%[language=JSON]
 "_via_img_metadata": {
    "00.jpg353386": {
      "filename": "00.jpg",
      "regions": [
        {
          "shape_attributes": {"name": "rect", "x": 397, "y": 211, "width": 86, "height": 29},
          "region_attributes": {"field_name": "name", "value": "Refik", "features": {"lowercase": true}}
        },
      etc],
    }
\end{lstlisting}

For the fake Travel and ID Documents we added  the following information:

\begin{itemize}
    \item name: The new name of the file.
    \item ctype : The transformation that has been done to generate this new document.
    \item loader: Where the image came from (by default Midv2020).
    \item src: The relative path of the template image used to create the new document.
    \item second src: The relative path of a second template image if it has been needed. For example in the case of the {\em crop \& replace}.
    \item field: Which field has been modified.
\end{itemize}

The code below illustrate and example of the metadata created for the fake documents.

\begin{lstlisting}[language=Python]
        name:str
        ctype:str ["Crop_and_replace", "Inpaint_and_rewrite"]
        loader:str ["Dataset where the image comes from"]
        src:str ["Src Image"]
        second_src:str ["Other Image in case CR"]
        field:str ["field modified"]
        second_field:str ["second field modified in case CR"] 
\end{lstlisting}

Finally, for the extracted clips from the recorded videos of fake travel and ID Documents, we added to the metadata information the bounding box coordinates of the document location within  the counterfeit clip. We follow the same bounding box representation used for the original MIDV2020 dataset, as shown below:

\begin{lstlisting}[language=Python]
{
    "filename": str,
    "regions": [{"shape_attributes": {
                "name": "polygon",
                "all_points_x": [682,693,142,138],
                "all_points_y": [188,1103,1068,237]
            },
            "region_attributes": {
                "field_name": "doc_quad"}
                }
}
\end{lstlisting}


\section*{Technical Validation}

Data is partitioned to allow three model validation techniques: {\bf hold-out},  {\bf k-fold cross-validation} and {\bf few-shot}. The code provided allow users to define any partition for these three  model validation schemes. Data is randomly sampled and it is, by default, split into 80\%-10\%-10\% for the hold-out validation and split it into 10 folds for the k-fold cross-validation. For few-shot, 6 nationalities ID documents are randomly chosen for the meta-training and the remainder 4 for the meta-testing. However, for fair comparison between models we also provide a predefined partition in which training, validation and test instances are always the same for both, the hold-out and the k-fold cross-validation.

To evaluate the SIDTD dataset, we have trained five state-of-the-art deep learning models for three tasks: {\bf template-based ID documents},  {\bf video-based ID documents} and {\bf Few-shot}. The template-based ID documents tasks evaluate the performance of the selected models to detect forged document that have solely altered by  {\em Composite} PAIs. The video-based ID document task evaluate the performance of the same models when the images comes from a video recording. The few-shot task is a quite more realistic task as models are trained for a given subset of nationalities and test on ID documents from other nationalities. Finally, to compare the dataset difficulty we compare the performance of the benchmark models on a private, and not available, dataset.

\subsection*{Deep Learning models}

The following five models were used to evaluate the SIDTD dataset: EfficientNet-B3~\cite{tan2019efficientnet}, ResNet50~\cite{he2016deep}, Vision Transformer Large Patch 16 (ViT-L/16)~\cite{dosovitskiy2020vit}, TransFG~\cite{he2022transfg} and Co-Attention Attentive Recurrent Network (CoAARC)~\cite{Berenguel:2019,wu2020acr}. 
EfficientNet, ResNet, and ViT models are general purpose models while TransFG is a model for fine-grained classification task while the CoAARC  model was designed to detect forged ID documents.

The EfficientNet-B3 and  ResNet-50 models are convolutional models widely used by the deep learning community for general purpose classification tasks. EfficientNet is a network model designed to use its  parameters more efficiently, optimising both accuracy and FLOPS\footnote{FLOPS = Floating point operations per second}, which has resulted in models with fewer parameters but equally accurate. ViT-L/16 and TransFG are models inspired by the Transformer\cite{vaswani2017attention} encoder architecture from NLP models. The Vision Transformer (ViT) model links each patch to the classification token through a self-attention mechanism based on a Transformer encoder. The TransFG models is an extension of the ViT model. Its main innovation is the addition of a Part Selection Module that selects the tokens that bring valuable information in order to use only discriminative image patches. We use the same type of ViT model for TransFG as a base network: ViT-L/16. 
The CoAARC model  is an algorithmic imitation of the human way that compares alternatively two images. The model contains a co-attention mechanism that focuses on identifying the most relevant and crucial parts of the images. For the given PAD task it focus on learning the differences between {\em bona fide} and forged ID Documents.

The EfficientNet-B3, ResNet50 and ViT-L/16 are built-in models from PyTorch\footnote{PyTorch site: https://pytorch.org} packages. ViT and CoAARC are trained with an input image resolution of $224x224x3$, and EfficientNet-B3, ResNet50 and TransFG with image resolution of 299x299x3. Each model is pretrained on the ImageNet\cite{deng2009imagenet} dataset. 

\subsection*{Private Dataset}

To compare the  SIDTD dataset  to real, but private, industry datasets, we evaluate the performance of the above-mentioned models on a industry dataset composed of real-life images from the \emph{IDNow}\footnote{https://www.idnow.io/} production flow. These images are captured using various devices (scan, smartphone) without any constraint.  As in a real-world dataset there are much less forged ID documents than {\em bona fide} ones, the sub-set of forged ID documents is mainly created by the following {\em Composite} PAI techniques: 
\begin{itemize}
    \item \textbf{Copy \& Paste} operations inside the same document. Similar to the {\em Crop \& Replace} technique described above, the personal data fields from a real document are firstly located. Then, a source and a target fields are randomly selected and the content of the source field is copied and pasted in the target field.
    \item \textbf{Copy \& Move} operations. Similar to the {\em Crop \& Replace} technique described above, a forged document  is created from a real document by replacing its original identity photo with another one randomly selected from a given set of identity photos. 
\end{itemize}

The data set of this private dataset is finally composed of $1,000$ {\em bona fide} examples and their corresponding $1,000$ {\em forged} examples. All the documents of this data set, whether {\em bona fide} or forged, are of various types (identity car, passports, driving licence, resident permit, etc.) from different countries (France, Spain, Italy, Romania, etc.).

\subsection*{Results}

Most of the models performs well in the two supervised tasks on the SIDTD and private dataset. The challenge of verifying if an ID or travel document image comes from an original document or it has been modified, or forged, by any means is relatively easy if models are trained with enough data from a given set of the nationalities. The problem comes when models try to verify whether ID or travel documents of unknown nationalities have been forged or not. The results for the Few-shot task support this thesis. Models performance drop down for all them in both metrics: accuracy and ROC AUC. The same behavior is observed on the private datasets. Model performance are overall good but performance drops on a few-shot scenario.

\begin{table}[h]
\centering
\resizebox{\textwidth}{!}{
\begin{tabular}{|c|c|c|c|c|c|c|} 
\hline
 \multirow{2}{*}{Dataset} &  \multicolumn{2}{c|}{Templates-based task }  & \multicolumn{2}{c|}{Video-based task}  &   \multicolumn{2}{c|}{ Few-shot  task}   \\   \cline{2-7} 
  & accuracy & ROC AUC & accuracy & ROC AUC & accuracy & ROC AUC  \\
\hline 
EfficientNet & $0.994 \pm 0.010$ & $1.000 \pm 0.000$ & $0.999 \pm 0.001$ & $1.000 \pm 0.000$   &  $0.593 \pm 0.023$ & $0.623 \pm 0.036$  \\  
ResNet & $0.981 \pm 0.012$ & $1.000 \pm 0.011$ & $0.999 \pm 0.002$ & $1.000 \pm 0.000$  &  $0.635 \pm 0.058$ & $0.669 \pm 0.073$  \\ 
\hline 
ViT & $0.552 \pm 0.023$ & $0.501 \pm 0.042$ & $0.975 \pm 0.020$ & $0.989 \pm 0.015$ &  $0.609 \pm 0.026$ & $0.645 \pm 0.032$  \\  
TransFG & $0.966 \pm 0.015$ & $0.992 \pm 0.004$ & $0.999 \pm 0.002$  & $1.000 \pm 0.000$ & $0.582 \pm 0.038$ & $0.612 \pm 0.049$  \\ 
\hline 
CoAARN & $0.986 \pm 0.016$ & $0.999 \pm 0.003$ & $0.992 \pm 0.012$ & $0.999 \pm 0.003$ &  $0.530 \pm 0.020$ & $0.539  \pm 0.027$  \\ 
\hline 
\end{tabular} 
}
\caption{Models performance in  terms of accuracy and ROC AUC  scores with standard deviation. Models are trained and test regarding a 10-fold cross-validation for the template-based ID document and the Video-based ID document tasks.  }
\label{tab:all_results}
\end{table}

\input{usage_notes}

%

\input{code_availability}

\bibliographystyle{unsrt}  
\bibliography{sample}


\section*{Acknowledgements} 

SOTERIA has received funding from the European Union’s Horizon 2020 	research and innovation programme under grant agreement No 101018342. 

\end{document}

%% file: usage_notes.tex
\section*{Usage Notes}

%



The main functionalities of the dataset are divided into two sections: (i) Loading the dataset and  (ii) generating a new dataset. All the necessary steps to use the dataset and generate new samples are described in the code repository.

\subsection*{DataLoader}
The dataloader is prepared to download the full dataset, split it into the validation partitions and load it in the system memory. For instance, to download the dataset and use the template instances of the ID Documents the code below have to be copied in a Python script: %
\begin{lstlisting}[language=Python]
from SIDTD.data.DataLoader.Datasets import *

data = SIDTD(download_original=False, custom_path_to_download=None).download_dataset("templates")
\end{lstlisting}

Alternatively, data can directly downloaded from a shell terminal, with the proper Python configuration as shown below: %
\begin{lstlisting}[language=Python]
python data_loader.py [--dataset DATASET] [--kind KIND] [--download_static] [--type_split TYPE_SPLIT] [--unbalanced] [-c|--cropped]
\end{lstlisting}

The optional parameter \texttt{--download\_static} is configured to use predefined data partitions, ensuring a fair comparison with other methods. Alternatively, if not set, random data partitions are generated based on the \texttt{--type\_split} parameter.

\subsection*{DataGeneration}

We also provide the functionality  to generate more images using the techniques described in the Methods section. The following example generate new fake data, based on the MIDV2020 dataset: 

\begin{lstlisting}[language=Python]
python data/explore/generate_fake_dataset.py  -dt Midv2020 -s ./SIDTD/templates
\end{lstlisting}

Within the repository, there is a subfolder located in the \textit{data} directory, named \textit{explore}. This folder contains  code examples showcasing the functions used for the creation of counterfeit ID.

\subsection*{Models}

The code is also ready to use the trained models and generate the csv files used to compute the results shown in Table~\ref{tab:all_results}. See below an example to reproduce the classification results with EfficientNet model (with and without GPU) on template instances of the ID documents:

\begin{lstlisting}[language=Python]
# Test EfficientNet model with CUDA
python test.py --name='EfficientNet' --dataset='SIDTD' --model='efficientnet-b3'

# Test EfficientNet model with CPU
python test.py --name='EfficientNet' --dataset='SIDTD' --model='efficientnet-b3' --device='cpu'
\end{lstlisting}


%% file: code_availability.tex
\section*{Code availability}


The code developed to download the data and prepare it to be used for models training and testing is available at the public code repository \hyperlink{https://github.com/Oriolrt/SIDTD\_Dataset}{https://github.com/Oriolrt/SIDTD\_Dataset}. Every model is coded in Pytorch.  All the scripts are coded with Python 3.7> and the setup.py is ready  to install all the package dependencies. We strongly recommend to use Python environments to avoid package version dependencies issues. 
The models used to report results on the SIDTD dataset can be downloaded from the CVC repository. 

%% file: sidtd_paper.bbl
\begin{thebibliography}{10}

\bibitem{IOCTA2020}
Catherine De~Bolle and et~al.
\newblock Internet organised crime thread assesment (iocta).
\newblock {\em EUROPOL}, 2020.

\bibitem{Bulatov2021}
Konstantin~B. Bulatov, Ekaterina Emelianova, Daniil~V. Tropin, Natalya
  Skoryukina, Yulia~S. Chernyshova, Alexander Sheshkus, Sergey~A. Usilin,
  Zuheng Ming, Jean{-}Christophe Burie, Muhammad~Muzzamil Luqman, and
  Vladimir~V. Arlazarov.
\newblock {MIDV-2020:} {A} comprehensive benchmark dataset for identity
  document analysis.
\newblock {\em CoRR}, abs/2107.00396, 2021.

\bibitem{Benalcazar2022}
Daniel Benalcazar, Juan~E. Tapia, Sebastian Gonzalez, and Christoph Busch.
\newblock Synthetic id card image generation for improving presentation attack
  detection, 2022.

\bibitem{iso3017}
Information technology - biometric presentation attack detection - part 1:
  Framework.
\newblock Technical report.

\bibitem{chazalon2017smartdoc}
Joseph Chazalon, Petra Gomez-Kr{\"a}mer, J-C Burie, Micka{\"e}l Coustaty,
  S{\'e}bastien Eskenazi, M~Luqman, Nibal Nayef, Mar{\c{c}}al Rusi{\~n}ol,
  Nicolas Sidere, and J-M Ogier.
\newblock Smartdoc 2017 video capture: Mobile document acquisition in video
  mode.
\newblock In {\em 2017 14th IAPR International Conference on Document Analysis
  and Recognition (ICDAR)}, volume~4, pages 11--16. IEEE, 2017.

\bibitem{karras2019style}
Tero Karras, Samuli Laine, and Timo Aila.
\newblock A style-based generator architecture for generative adversarial
  networks.
\newblock In {\em Proceedings of the IEEE/CVF conference on computer vision and
  pattern recognition}, pages 4401--4410, 2019.

\bibitem{tan2019efficientnet}
Mingxing Tan and Quoc Le.
\newblock {Efficientnet}: {R}ethinking model scaling for convolutional neural
  networks.
\newblock In {\em International conference on machine learning}, pages
  6105--6114.

\bibitem{he2016deep}
Kaiming He, Xiangyu Zhang, Shaoqing Ren, and Jian Sun.
\newblock Deep residual learning for image recognition.
\newblock In {\em Proceedings of the IEEE conference on computer vision and
  pattern recognition}, pages 770--778, 2016.

\bibitem{dosovitskiy2020vit}
Alexey Dosovitskiy, Lucas Beyer, Alexander Kolesnikov, Dirk Weissenborn,
  Xiaohua Zhai, Thomas Unterthiner, Mostafa Dehghani, Matthias Minderer, Georg
  Heigold, Sylvain Gelly, et~al.
\newblock An image is worth 16x16 words: Transformers for image recognition at
  scale.
\newblock {\em arXiv preprint arXiv:2010.11929}, 2020.

\bibitem{he2022transfg}
Ju~He, Jie-Neng Chen, Shuai Liu, Adam Kortylewski, Cheng Yang, Yutong Bai, and
  Changhu Wang.
\newblock Trans{FG}: A transformer architecture for fine-grained recognition.
\newblock In {\em Proceedings of the AAAI Conference on Artificial
  Intelligence}, volume~36, pages 852--860, 2022.

\bibitem{Berenguel:2019}
Albert Berenguel~Centeno, Oriol Ramos~Terrades, Josep Llad{\'{o}}s, and
  Cristina Ca{\~{n}}ero.
\newblock Recurrent comparator with attention models to detect counterfeit
  documents.
\newblock In {\em 2019 International Conference on Document Analysis and
  Recognition, {ICDAR} 2019, Sydney, Australia, September 20-25, 2019}, pages
  1332--1337. {IEEE}, 2019.

\bibitem{wu2020acr}
Lin Wu, Yang Wang, Junbin Gao, Meng Wang, Zheng-Jun Zha, and Dacheng Tao.
\newblock Deep coattention-based comparator for relative representation
  learning in person re-identification.
\newblock {\em IEEE transactions on neural networks and learning systems},
  32(2):722--735, 2020.

\bibitem{vaswani2017attention}
Ashish Vaswani, Noam Shazeer, Niki Parmar, Jakob Uszkoreit, Llion Jones,
  Aidan~N Gomez, {\L}ukasz Kaiser, and Illia Polosukhin.
\newblock Attention is all you need.
\newblock {\em Advances in neural information processing systems}, 30, 2017.

\bibitem{deng2009imagenet}
Jia Deng, Wei Dong, Richard Socher, Li-Jia Li, Kai Li, and Li~Fei-Fei.
\newblock Imagenet: A large-scale hierarchical image database.
\newblock In {\em 2009 IEEE conference on computer vision and pattern
  recognition}, pages 248--255. Ieee, 2009.

\end{thebibliography}
